# CataractSAM-2: A Domain-Adapted Model for Anterior Segment Surgery Segmentation and Scalable Ground-Truth Annotation

Mohammad Eslami[1,*], Dhanvinkumar Ganeshkumar[2,*], Saber Kazeminasab[1], Michael G. Morley[3], Michael V. Boland[3], Michael M. Lin[3], John B. Miller[3], David S. Friedman[3], Nazlee Zebardast[3], Lucia Sobrin[3], and Tobias Elze[1]

[1]Harvard Ophthalmology AI Lab, Massachusetts Eye and Ear, Harvard Medical School, Boston, MA, USA
[2]Thomas Jefferson High School for Science and Technology, Chantilly, Virginia, USA
[3]Massachusetts Eye and Ear, Harvard Medical School, Boston, MA, USA
[*]Equal Contribution

## Abstract

We present CataractSAM-2, a domain-adapted extension of Meta's Segment Anything Model 2, designed for real-time semantic segmentation of cataract ophthalmic surgery videos with high accuracy. Positioned at the intersection of computer vision and medical robotics, CataractSAM-2 enables precise intraoperative perception crucial for robotic-assisted and computer-guided surgical systems. Furthermore, to alleviate the burden of manual labeling, we introduce an interactive annotation framework that combines sparse prompts with video-based mask propagation. This tool significantly reduces annotation time and facilitates the scalable creation of high-quality ground-truth masks, accelerating dataset development for ocular anterior segment surgeries. We also demonstrate the model's strong zero-shot generalization to glaucoma trabeculectomy procedures, confirming its cross-procedural utility and potential for broader surgical applications. The trained model and annotation toolkit are released as open-source resources, establishing CataractSAM-2 as a foundation for expanding anterior ophthalmic surgical datasets and advancing real-time AI-driven solutions in medical robotics, as well as surgical video understanding.

## 1 Introduction

Real-time semantic segmentation is critical in robotic and computer-aided surgery, enabling precise localization of anatomical structures and surgical tools [Bouget et al., 2017, Rueckert et al., 2024, Varga and Poncelet, 2025]. This capability underpins key functions such as surgical navigation, intraoperative documentation, and robotic assistance [Ahmed et al., 2024]. Beyond these core applications, segmentation supports advanced use cases including augmented reality (AR) overlays and step recognition in cataract and vitreoretinal surgery [Tu et al., 2023, Yang and Kim, 2024, Zisimopoulos et al., 2018, Twinanda et al., 2017], virtual reality (VR) simulators for ophthalmic training, and early autonomous suturing systems such as the Smart Tissue Autonomous Robot (STAR), which demonstrated the feasibility of real-time vision-guided interventions [Leonard et al., 2014]. More recently, robust segmentation overlays have been shown to enhance capsulorhexis precision and reduce intraoperative error rates [Chen et al., 2025].

These advances are increasingly vital in robotic surgery, where automation, real-time perception, and reliable tool tracking are prerequisites for safe and precise interventions. Nonetheless, significant challenges persist due to transparent tissues, glare, and the visual similarity of surgical instruments [Ghamsarian et al., 2021].

While temporal foundation models like Meta's Segment Anything Model 2 (SAM-2) [Ravi et al., 2024] show promise for general-purpose segmentation, their zero-shot performance varies significantly when applied to surgical video without domain adaptation [Mazurowski et al., 2023]. Prior extensions such as MedSAM2 [Zhou et al., 2023] and SurgSAM2 [Zhang et al., 2024] demonstrate the importance of tailoring SAM-based models to medical and surgical contexts through dataset-specific fine-tuning and temporal optimization [Wu et al., 2023]. Recently, Zhaksylyk *et* *al.* introduced RP-SAM2 [Zhaksylyk et al., 2025], which augments SAM-2 with a lightweight shift-block to correct noisy prompts, improving single-click instrument segmentation accuracy on Cataract-1K. However, the ophthalmic domain remains relatively underexplored, despite its global clinical prevalence and unique visual complexity.

Despite these promising adaptations, progress remains bottlenecked by annotation costs. Currently, all available ophthalmic surgery video datasets for segmentation or detection focus exclusively on cataract procedures. A key bottleneck in adapting segmentation models for surgery is the cost of generating ground-truth masks. Creating ground-truth segmentation masks for surgical videos is labor-intensive, often requiring expert clinicians to label frames in detail. A single intraoperative video or volumetric scan can take many hours to annotate manually [Marinov et al., 2023, Ramadan et al., 2020]. There is a growing consensus that more intuitive, human-in-the-loop annotation tools are needed to boost efficiency.

Advances in surgical AI and medical robotics increasingly rely on real-time perception, precise tool tracking, and scalable dataset generation. Building on these needs, our key contributions are as follows:

1. *CataractSAM-2*: A domain-adapted variant of SAM-2, specifically fine-tuned for real-time, high-precision segmentation in cataract and anterior segment ophthalmic surgeries.
2. *Interactive Annotation Framework*: A human-in-the-loop annotation pipeline built upon SAM-2's video predictor, allowing users to provide sparse prompts (e.g., a single point on an object in one frame) via an intuitive user interface (UI) embedded in Jupyter cells. The framework automatically propagates high-quality masks across all frames, significantly reducing annotation time and enabling the rapid development of labeled surgical datasets.
3. *External Validation on CaDIS*: We externally validated CataractSAM-2 on the CaDIS dataset, the largest publicly available cataract pixel-wise segmentation benchmark. The model achieved high accuracy and consistent performance across diverse cases, validating transfer to additional instruments and surgical phases beyond Cataract-1K.
4. *Cross-Procedure Generalization*: We demonstrate that CataractSAM-2 generalizes beyond cataract surgery by successfully segmenting surgical scenes in trabeculectomy glaucoma videos—procedures not seen during training—highlighting its robustness across anterior segment surgeries.
5. *Open-Source Release*: To foster reproducibility and accelerate downstream research, we release all pretrained model weights, inference scripts, and the interactive annotation notebook via (GitHub)(GitHub/backup) and (Hugging Face).
6. *Impact on Dataset Generation*: All above contributions together enable scalable dataset creation for anterior segment surgeries and help advance the field of surgical AI and medical robotics.

## 2 Data and Methodolgy

### 2.1 Data

We used the publicly available Cataract-1K dataset, consisting of 30 cataract surgery videos (2,256 annotated frames) recorded between 2021 and 2023 at the Klinikum Klagenfurt Eye Clinic. Each video was captured at a resolution of 1920 × 1080 pixels and 25 frames per second (FPS) [Ghamsarian et al., 2024]. Clinical experts provided annotations for key anatomical structures and surgical instruments, as detailed explicitly in Table 1. Ground-truth binary segmentation masks were provided for each annotated frame and served as a benchmark for evaluating the model. An 80/20 train-test split was applied, with 24 videos used for training and 6 for testing.

Table 1: Annotated Classes in the Cataract-1K Dataset

| Category | Annotated Classes |
|---|---|
| Anatomical Structures | • Iris • Pupil • Lens |
| Surgical Instruments | • Slit/Incision Knife • Gauge Sizes • Capsulorhexis Cystotome • Capsulorhexis Forceps • Katena Forceps • Phacoemulsifier Tip • Spatula • Irrigation-Aspiration • Lens Injector |

Furthermore, for external evaluation, we used the CaDIS dataset as the test set for CataractSAM-2 trained on Cataract-1K. Also, to assess the transferability of our model to other anterior segment procedures, we qualitatively evaluated CataractSAM-2 on two publicly available videos (YouTube) of glaucoma trabeculectomy surgery, which were not used during training, to serve as test cases for evaluating the generalizability:

- **Video GL1:** *Trabeculectomy for Treatment of Glaucoma – Edited Surgery Pearls* [of South Texas, 2021].

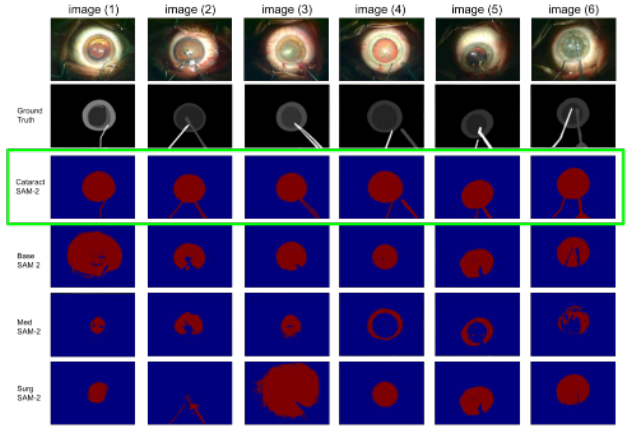

Figure 1: Random examples of binary segmentation predictions from SAM-2 models using SAM2ImagePredictor, shown alongside test frames and ground truth masks; performance differences are expected as only our model was trained on cataract surgery.

- **Video GL2:** *Trabeculectomy – Glaucoma Surgery with Mitomycin-C* [Hospital, 2020].

## 2.2 Segmentation Models

This study evaluates the performance of four deep learning models for the semantic segmentation of surgical videos, focusing on cataract and glaucoma procedures (RP-SAM2 is not publicly available at this time):

**Base SAM-2**: The Segment Anything Model 2 (SAM-2) is a vision transformer-based foundation model designed by Meta for general-purpose image segmentation. Its architecture comprises three modules: an image encoder to extract visual features, a prompt encoder to process user-defined inputs (e.g., points, boxes, or masks), and a mask decoder that generates segmentation outputs [Ravi et al., 2024]. This design enables prompt-based segmentation, although latency and domain mismatch limit its utility in surgical videos.

**MedSAM-2**: MedSAM-2 extends SAM-2 to static 2D and 3D medical images by reframing segmentation as a video tracking task, leveraging temporal continuity to enhance spatial coherence across slices [Zhou et al., 2023].

**SurgSAM-2**: SurgSAM-2 is a real-time surgical segmentation framework that adapts SAM-2 for laparoscopic and endoscopic procedures. It introduces an Efficient Frame Pruning (EFP) module that computes cosine similarity between sequential video frames to eliminate redundancy and improve processing speed [Zhang et al., 2024].

**CataractSAM-2 (Ours)**: CataractSAM-2 is our domain-adapted extension of Meta's Segment Anything Model (SAM-2), trained specifically for real-time segmentation of ophthalmic surgery videos. The model retains SAM-2's original vision transformer-based *image encoder* to preserve general visual priors, while fine-tuning the *prompt encoder* and *mask decoder* to handle the unique spatial complexity and visual challenges of cataract procedures. CataractSAM-2 is optimized for segmenting key anatomical structures (e.g., cornea, iris, lens capsule) and surgical tools (e.g., phacoemulsifier probe, irrigation cannula, etc.), which are often difficult to detect due to low contrasts and occlusion blur in surgical footage. The following training configuration was used to balance accuracy and efficiency: AdamW optimizer (learning rate = 0.0001, weight decay = 1e-4); mixed-precision training with PyTorch AMP; gradient accumulation every four steps to simulate larger batch sizes on limited hardware; step-based learning rate scheduler with decay every 500 steps; and a warm-up phase on the first 5 frames of each video to stabilize early predictions.

## 3 Results

### 3.1 Evaluation Strategy

We evaluated segmentation performance using the ground-truth masks provided in Cataract-1K, ensuring consistent comparison across all models [Reinke et al., 2024, Yu et al., 2022]. An 80/20 train-test split was applied, with 24 videos used for training and 6 for testing. For both training and testing, we use binary masks in which all annotated anatomical structures and surgical tools (e.g., cornea, iris, lens capsule, phacoemulsifier probe) are merged into a single foreground class. Each binary mask assigns 1 to the foreground and 0 to the background.

Segmentation performance is reported for each test video individually. For each video, we compute the mean and standard deviation of Intersection over Union (IoU), and Pixel Accuracy (PAC) across its frames to assess intra-video consistency and performance. *IoU:* Measures the overlap between predicted masks and ground truth. It is the primary metric for region-based segmentation accuracy. IoU is stricter than Dice because it penalizes false positives and false negatives more heavily, since denominator is the union. *PAC:* Calculates the proportion of correctly classified pixels across the entire frame, providing a frame-level performance overview.

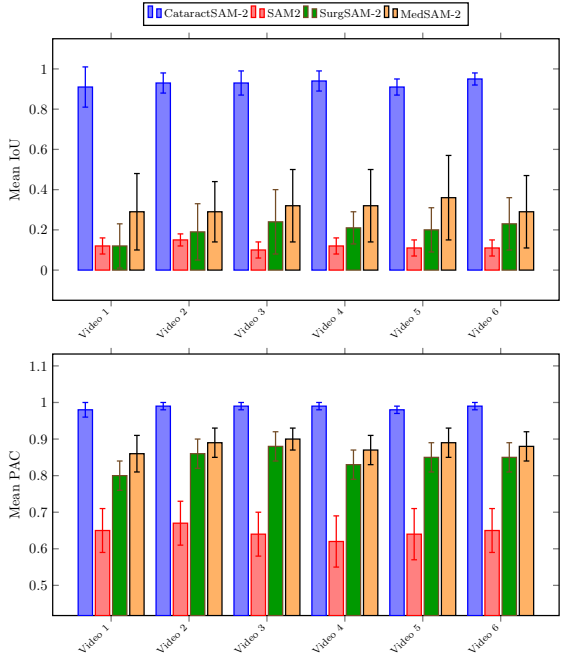

Figure 2: Performance Comparison for IoU, and PAC Across Six Test Videos.

## 3.2 Results

We evaluated CataractSAM-2, SAM-2, MedSAM-2, and SurgSAM-2 on Cataract-1K and trabeculectomy videos to assess segmentation performance and cross-procedural generalizability through visual, quantitative, and cross-domain comparisons.

### 3.2.1 Qualitative Comparison

We present visual comparisons of binary segmentation masks generated by four SAM-2 variants on six representative frames from the six test videos from Cataract-1K. As shown in Figure 1, each column corresponds to a distinct surgical frame, with rows displaying the original image, ground truth mask, and predictions from CataractSAM-2, base SAM-2, MedSAM-2, and SurgSAM-2. As expected since CataractSAM-2 is the only one fine-tuned for cataract surgery, it consistently produces clean, well-defined masks that closely align with the ground truth, even under challenging conditions such as occlusion and glare.

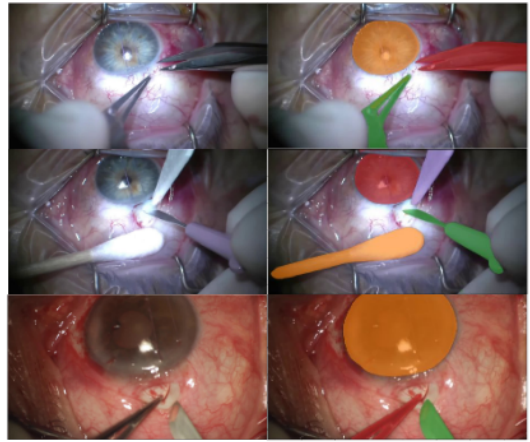

Figure 3: CataractSAM-2 generalization to trabeculectomy surgery using SAM2VideoPredictor. From top to bottom: (1) initial conjunctival incision, (2) scleral flap dissection, and (3) scleral flap creation during trabeculectomy. Left: original video frame. Right: predicted segmentation by CataractSAM-2. More frames are shown in Appendix.

### 3.2.2 Quantitative Comparison

CataractSAM-2 achieves real-time performance with an inference speed of 15 frames per second (FPS) in binary segmentation, making it suitable for intraoperative deployment *(NVIDIA A100 GPU)*. To benchmark segmentation quality, we evaluated all models on the six test videos from Cataract-1K using the metrics and shown the results in Figure 2. Across all six cataract surgery videos, CataractSAM-2 consistently outperforms baseline models on all metrics: IoU and PAC. SurgSAM-2 and MedSAM-2 show moderate improvements, particularly in pixel-level accuracy, but still struggle with tool boundaries and exhibit high variability across videos. In contrast, CataractSAM-2 achieves both high accuracy and low variability, especially in delineating tool boundaries and anatomical regions. These results highlight its robustness under challenging intraoperative conditions.

### 3.2.3 External Dataset Evaluation

To further validate the robustness of CataractSAM-2, we evaluated the performance of the model trained on Cataract-1K using the CaDIS dataset, which contains 25 full-length cataract surgery videos (4,613 frames) with 30 annotated instrument and anatomical classes, compared to the 12 classes in Cataract-1K [Grammatikopoulou et al., 2021]. For consistency, only the 12 classes common to both datasets were considered.

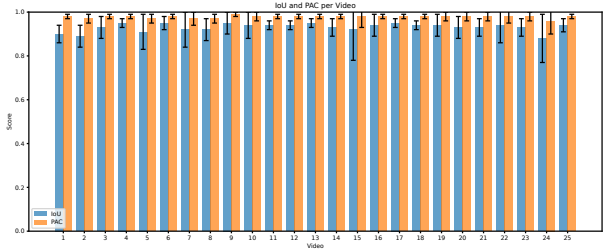

Figure 4: CataractSAM-2 performance on the CaDIS dataset across 25 cataract videos. Metrics are reported as mean ± standard deviation.

Figure 4 reports the per-video mean and standard deviation for PAC and IoU.

Across the 25 CaDIS videos, performance was consistent with IoU ranging from 0.88–0.95, and PAC from 0.96–0.99. The results highlight the model's ability to generalize beyond the Cataract-1K training domain to a larger, heterogeneous set of instruments and surgical phases comparable to state-of-the-art [Zhang et al., 2025].

#### 3.2.4 Qualitative Visualization – Transferability

This task challenges the model to generalize across other anterior segment surgeries and differing anatomical structures, surgical tools, and visual conditions. Figure 3 presents qualitative segmentation results from CataractSAM-2 across three representative trabeculectomy scenes, each corresponding to a different surgical phase: (1) initial incision, (2) scleral flap dissection, and (3) scleral flap creation. In the initial incision phase (top row), CataractSAM-2 accurately segments the iris and both instruments (a forceps and a blade), despite low contrast and close spatial proximity near the pupil. The mask boundaries align well with tool tips and anatomical features. During scleral flap dissection (middle row), the model segments three instruments: Weck-Cel (cellulose spear) at the top right, a slit knife at the bottom right, and a cotton-tip applicator at the bottom left, even under glare and overlapping geometry. While overall shape consistency is maintained, some masks slightly exceed object boundaries, indicating minor oversegmentation in high-reflectivity zones. In the scleral flap creation phase (bottom row), CataractSAM-2 segments both forceps (bottom left) and a slit knife (bottom right), maintaining clear separation from surrounding tissues despite partial occlusion and irregular geometry. The model maintains clear separation from surrounding tissue structures. These results demonstrate that CataractSAM-2 generalizes effectively to unseen ophthalmic procedures, preserving segmentation fidelity across varied surgical scenes, consistent with recent auto-reprompting frameworks that improve temporal coverage in surgical video [Sivakumar et al., 2025]. Comparable cross-procedure trends appear in ophthalmic workflow datasets, where models trained on standard cataract techniques retain approximately 85–89% phase-recognition accuracy on small-incision cataract surgery [Mueller et al., 2025].

## 4 Interactive Ground-Truth Annotation Framework

Overall, the results presented in the previous section indicate that **CataractSAM-2 is an end-to-end foundation model for cataract segmentation and can substantially accelerate anterior segment annotation workflows**.

To support scalable dataset creation beyond cataract surgery and address the bottleneck of manual annotation in anterior ophthalmic surgical video segmentation, we developed an interactive demo notebook based on CataractSAM-2. This tool enables users to generate high-quality segmentation masks with minimal input through an intuitive interface, leveraging SAM's prompt-based capabilities and our domain-adapted video predictor. The system achieves a segmentation speed of approximately 4 frames per second (FPS) on a standard GPU, enabling efficient mask propagation across surgical videos with low latency.

Interactive methods can reduce expert labeling time by an order of magnitud, down from about an hour to ∼3 minutes per volume via iterative feedback [Marinov et al., 2023]. In video, pairing SAM with tracking enables one-pass propagation from sparse prompts [Yang et al., 2023]. Moreover, correlation-aware active learning selects a diverse ∼30% of frames per clip with no accuracy loss, further cutting annotation burden [Wu et al., 2024].

Figure 5 illustrates the interactive segmentation pipeline:

- **Step 1:** Multiple objects can be annotated in the first frame using positive and negative points via the `Object()` function. Specifically, `Object(frame#, classID, className)` initializes a new interactive segmentation instance at the specified frame and assigns a unique identifier to the object. Instead of manually entering input prompts, users can visually select *positive*

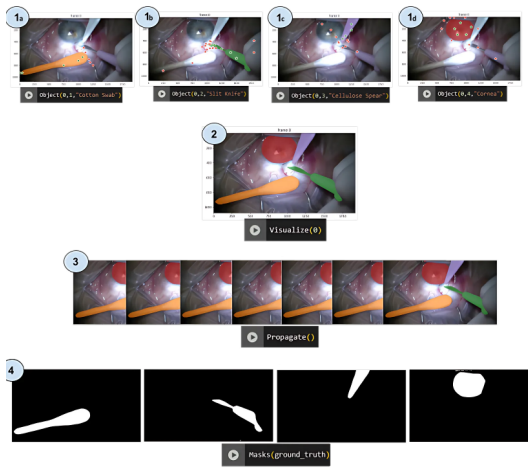
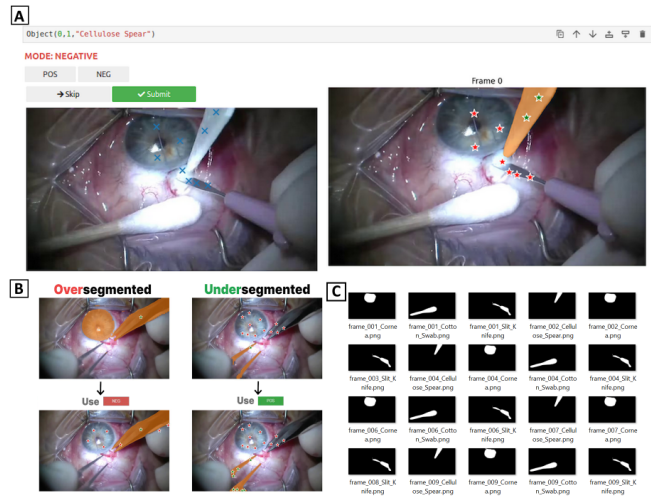

The annotation pipeline consists of the following steps: 1) Prompt-based annotation (see subfigure A), 2) Visualization to review all masks, 3) Propagation of masks across all frames using SAM2VideoPredictor, and 4) Saving the final masks.

A) In-cell UI for object annotation using prompt-based positive and negative points. B) Corrections of oversegmentation (left) and undersegmentation (right) using point prompts. C) Generated masks for each frame and object.

Figure 5: The interactive prompt-based annotation pipeline, user interface, and the resulting masks for a four-class instrument video clip. A video demonstration is available at https: (removed for anonymity, but available in supplementary material)

points (to correct undersegmentation) and *negative* points (to correct oversegmentation) directly on the frame (see Figure 5b). The annotator may label as many frames and object classes as needed.

- **Step 2:** The `Visualize()` function renders the fully segmented frame with all annotated objects for visual verification. If a segment is not good enough, user can edit its annotation by rerunning the related cell and annotating again.

- **Step 3:** The `Propagate()` function automatically tracks and segments the selected objects across subsequent frames using temporal consistency.

- **Step 4:** The `Masks(ground_truth)` function extracts each annotated object from the propagated outputs and saves them as a separate binary `.png` file.

- **(Optional) Step 5:** If the results are not satisfactory, the user can use the `Restart()` function to re-initialize points.

## 5 Limitations and Future Work

While CataractSAM-2 demonstrates strong performance on ophthalmic surgical segmentation tasks, several limitations remain. First, dataset diversity is constrained. Cataract-1K consists primarily of cataract surgeries from a single clinical sites, reducing exposure to variability in imaging conditions, surgical styles, and patient anatomy. Multi-center studies demonstrate 5–20 point IoU reductions when models are applied across surgical procedures, underscoring the importance of cross-site validation [Roß et al., 2021]. Beyond ophthalmology, multi-domain instrument datasets such as Kvasir-Instrument, RoboTool, and dVRK provide broader benchmarks for tool segmentation and tracking [Jha et al., 2021, García-Peraza-Herrera et al., 2021, Colleoni et al., 2020, Psychogyios et al., 2023]. Although we evaluate zero-shot performance on trabeculectomy procedures, broader datasets across anterior segment surgeries are needed to assess generalization. Accuracy can be improved with data-centric adaptations—such as background synthesis [Wang et al., 2022] and one-to-many augmentation [Wang et al., 2023].

Second, the image encoder remains frozen during fine-tuning, limiting adaptation to ophthalmic-specific features such as glare and transparent tissues.

Lightweight tuning of encoder layers or pretraining on ophthalmic data may improve segmentation fidelity in visually complex scenarios. Specialized ophthalmic methods such adaptive tensor-based filtering for reliable pupil masks [Giap et al., 2024] and multi-view microscope fusion [Ou et al., 2022] suggest complementary avenues to further boost performance under glare/transparency.

Finally, although inference achieves near-real-time performance (15 FPS), latency remains a constraint for fully closed-loop surgical guidance or robotic systems. Model compression techniques, including pruning, quantization, and edge deployment (e.g., NVIDIA Jetson and Holoscan), should be explored to meet real-time constraints. As further future work, we plan to integrate step-aware active learning; prior work suggests target accuracy with ∼30% fewer labeled surgical videos [Shah et al., 2025].

# 6 Conclusion

We present CataractSAM-2, the first domain-adapted extension of Meta's Segment Anything Model 2 for anterior segment ophthalmic surgery segmentation. By fine-tuning only the prompt encoder and mask decoder on the Cataract-1K dataset, we achieve real-time, accurate performance. Furthermore, to reduce manual annotation, we introduce an interactive framework combining sparse prompts with video-based propagation. CataractSAM-2 achieves high segmentation accuracy on the CaDIS and generalizes well to trabeculectomy videos, demonstrating consistent performance across varying anatomy and conditions.

# References

[Ahmed et al., 2024] Ahmed, Y. et al. (2024). Robotic-assisted anterior segment procedures. *Ophthalmic Surgery Today*.

[Bouget et al., 2017] Bouget, D., Allan, M., Stoyanov, D., and Jannin, P. (2017). Vision-based and marker-less surgical tool detection and tracking: a review. *Medical Image Analysis*, 35:633–654.

[Chen et al., 2025] Chen, X., Liu, J., Liang, H., et al. (2025). Digitalization of surgical features improves surgical accuracy via surgeon guidance and robotization. *npj Digital Medicine*, 8(1):497.

[Colleoni et al., 2020] Colleoni, E., Flouty, E., Fawaz, N., Al-Mourad, M. B., and Stoyanov, D. (2020). dVRK ex vivo surgical tool segmentation dataset with kinematic data. Dataset.

[García-Peraza-Herrera et al., 2021] García-Peraza-Herrera, L., Fidon, L., D'Ettorre, E., and Vercauteren, T. (2021). RoboTool: An open multi-robot tool segmentation dataset. *arXiv preprint arXiv:2102.01766*.

[Ghamsarian et al., 2024] Ghamsarian, N., El-Shabrawi, Y., Nasirihaghighi, S., Putzgruber-Adamitsch, D., Zinkernagel, M., Wolf, S., Schoeffmann, K., and Sznitman, R. (2024). Cataract-1k dataset for deep-learning-assisted analysis of cataract surgery videos. *Scientific data*, 11(1):373.

[Ghamsarian et al., 2021] Ghamsarian, N., Taschwer, M., Putzgruber-Adamitsch, D., Sarny, S., El-Shabrawi, Y., and Schöffmann, K. (2021). Recal-net: Joint region-channel-wise calibrated network for semantic segmentation in cataract surgery videos. In *International Conference on Neural Information Processing*, pages 391–402. Springer.

[Giap et al., 2024] Giap, B. D., Srinivasan, K., Mahmoud, O., Mian, S. I., Tannen, B. L., and Nallasamy, N. (2024). Adaptive tensor-based feature extraction for pupil segmentation in cataract surgery. *IEEE Journal of Biomedical and Health Informatics*, 28(3):1599–1610.

[Grammatikopoulou et al., 2021] Grammatikopoulou, M., Flouty, E., Kadkhodamohammadi, A., Quelléc, G., Chow, A., Nehme, J., Luengo, I., and Stoyanov, D. (2021). CaDIS: Cataract dataset for surgical RGB-image segmentation. *Medical Image Analysis*, 71:102053. Free dataset: cataracts-semantic-segmentation2020 Grand Challenge.

[Hospital, 2020] Hospital, W. E. (2020). Trabeculectomy – glaucoma surgery with mitomycin-c. https://www.youtube.com/watch?v=csOvAMfOVdY. YouTube video.

[Jha et al., 2021] Jha, D., Ali, S., Gwinnutt, J., Valerio, D., Riegler, M. A., Halvorsen, P., and de Lange, T. (2021). Kvasir-instrument: Diagnostic and therapeutic instrument segmentation dataset in gastrointestinal endoscopy. *arXiv preprint arXiv:2101.05992*.

[Leonard et al., 2014] Leonard, S., Wu, K. L., Kim, P. C. W., and Taylor, R. H. (2014). Smart Tissue Anastomosis Robot (STAR): a vision-guided robotics system for laparoscopic suturing. In *Proc. IEEE Intl. Conf. on Robotics and Automation (ICRA)*, pages 2421–2427.

[Marinov et al., 2023] Marinov, Z., Jäger, P. F., Egger, J., Kleesiek, J., and Stiefelhagen, R. (2023). Deep interactive segmentation of medical images: A systematic review and taxonomy. *Medical Image Analysis*, 86:102806.

[Mazurowski et al., 2023] Mazurowski, M. A., Dong, H., Gu, H., Yang, J., Konz, N., and Zhang, Y. (2023). Segment Anything Model for medical image analysis: an experimental study. *Medical Image Analysis*, 89:102918.

[Mueller et al., 2025] Mueller, S., Sachdeva, B., Prasad, S. N., Wintergerst, M. W. M., Murali, K., Jain, M., Finger, R. P., Schultz, T., et al. (2025). Phase recognition in manual small-incision cataract surgery with MS-TCN++ on the novel SICS-105 dataset. *Scientific Reports*, 15(1):16886.

[of South Texas, 2021] of South Texas, E. A. (2021). Trabeculectomy for treatment of glaucoma – edited surgery pearls. https://www.youtube.com/watch?v=sIa-oOc0B-U. YouTube video.

[Ou et al., 2022] Ou, M., Li, H., Liu, H., Wang, X., Yi, C., Hao, L., Hu, Y., and Liu, J. (2022). MVD-Net: Semantic segmentation of cataract surgery using multi-view learning. In *Proc. IEEE EMBC*, pages 5035–5038.

[Psychogyios et al., 2023] Psychogyios, D., Flouty, E., Fawaz, N., and Stoyanov, D. (2023). SAR-RARP50: A comprehensive dataset for segmenting instruments in robotic radical prostatectomy. *Medical Image Analysis*, 87:102805.

[Ramadan et al., 2020] Ramadan, S., AlBadawy, E. A., Ismail, M., and El-Baz, A. (2020). HAL-IA: A hybrid active learning framework using interactive segmentation for medical image annotation. *Computerized Medical Imaging and Graphics*, 85:101782.

[Ravi et al., 2024] Ravi, N., Gabeur, V., Hu, Y.-T., Hu, R., Ryali, C., Ma, T., Khedr, H., Rädle, R., Rolland, C., Gustafson, L., Mintun, E., Pan, J., Alwala, K. V., Carion, N., Wu, C.-Y., Girshick, R., Dollár, P., and Feichtenhofer, C. (2024). Sam 2: Segment anything in images and videos. *arXiv preprint arXiv:2408.00714*.

[Reinke et al., 2024] Reinke, A., Tizabi, M. D., Pietsch, C., Unberath, M., et al. (2024). Metrics reloaded: recommendations for image analysis validation. *Nature Methods*.

[Roß et al., 2021] Roß, T., Reinke, A., Full, P. M., Wagner, M., Kenngott, H. G., Apitz, M., Hempe, H., and Maier-Hein, L. (2021). Comparative validation of multi-instance instrument segmentation in endoscopy: results of the ROBUST-MIS 2019 challenge. *Medical Image Analysis*, 70:101920.

[Rueckert et al., 2024] Rueckert, T., Rueckert, D., and Palm, C. (2024). Methods and datasets for segmentation of minimally invasive surgical instruments in endoscopic images and videos: a review of the state of the art. *Computers in Biology and Medicine*, 169:107929.

[Shah et al., 2025] Shah, N. A., Safaei, B., Sikder, S., Vedula, S. S., and Patel, V. M. (2025). StepAL: Step-aware active learning for cataract surgical videos. *arXiv preprint arXiv:2507.22059*.

[Sivakumar et al., 2025] Sivakumar, S. K., Frisch, Y., Ranem, A., and Mukhopadhyay, A. (2025). SASVi: Segment any surgical video. *International Journal of Computer Assisted Radiology and Surgery*, 20(8):1409–1419.

[Tu et al., 2023] Tu, R., Lin, Y., et al. (2023). Augmented reality in cataract surgery. *IEEE Transactions on Medical Imaging*.

[Twinanda et al., 2017] Twinanda, A. P., Shehata, S., Mutter, D., Marescaux, J., De Mathelin, M., and Padoy, N. (2017). EndoNet: A deep architecture for recognition of surgical workflow. *IEEE Transactions on Medical Imaging*, 36(1):252–265.

[Varga and Poncelet, 2025] Varga, B. and Poncelet, M. (2025). A shared control approach to robot-assisted cataract surgery training for novice surgeons. *Sensors*, 25(16):5165.

[Wang et al., 2022] Wang, A., Islam, M., Xu, M., and Ren, H. (2022). Rethinking surgical instrument segmentation: a background image can be all you need. In *Proc. MICCAI*, volume 13431 of *LNCS*, pages 355–364.

[Wang et al., 2023] Wang, A., Islam, M., Xu, M., and Ren, H. (2023). Generalizing surgical instruments segmentation to unseen domains with one-to-many synthesis. *arXiv preprint arXiv:2306.16285*.

[Wu et al., 2024] Wu, F., Marquez-Neila, P., Zheng, M., Rafi-Tari, H., and Sznitman, R. (2024). Correlation-aware active learning for surgery video segmentation. In *Proc. IEEE Winter Conference on Applications of Computer Vision (WACV)*, pages 2576–2585.

[Wu et al., 2023] Wu, J., Ji, W., Liu, Y., Fu, H., Xu, M., and Jin, Y. (2023). Medical SAM adapter: Adapting segment anything model for medical image segmentation. *arXiv preprint arXiv:2304.12620*.

[Yang and Kim, 2024] Yang, J. and Kim, D. (2024). Real-time navigation for vitreoretinal procedures. *Medical Image Analysis*.

[Yang et al., 2023] Yang, Z., Shen, Z., Wang, M., and Shao, L. (2023). Track Anything: Segment Anything meets videos. *arXiv preprint arXiv:2304.11968*.

[Yu et al., 2022] Yu, A. C., Mohajer, B., and Eng, J. (2022). External validation of deep learning algorithms for radiologic diagnosis: a systematic review. *Radiology: Artificial Intelligence*, 4(3):e210064.

[Zhaksylyk et al., 2025] Zhaksylyk, B., Li, S., Li, S., Chen, Y., Lu, Z., Yuan, Y., Gao, X., Shen, D., and Ma, K. (2025). Rp-sam2: Robust point-prompt segmentation in ophthalmic surgery. *arXiv preprint arXiv:2504.07117*.

[Zhang et al., 2025] Zhang, M., Gu, Y., Chen, X., Zheng, B., Wu, D., Zhang, J., Wu, Y., Liu, Y., and Zhao, Y. (2025). Enhancing trustworthiness of semantic segmentation in cataract surgery videos via intra-phase label propagation. *arXiv preprint arXiv:2504.07117*.

[Zhang et al., 2024] Zhang, W. et al. (2024). Surgsam2: Real-time segment anything in surgical video. *arXiv preprint arXiv:2408.07931*.

[Zhou et al., 2023] Zhou, T. et al. (2023). Medsam2: Segment anything in 3d medical images and videos. *arXiv preprint arXiv:2309.12957*.

[Zisimopoulos et al., 2018] Zisimopoulos, O., Flouty, E., Luengo, I., Stoyanov, D., and Deligianni, F. (2018). DeepPhase: Surgical phase recognition in CATARACTS videos. In *Proc. MICCAI*, volume 11071 of *LNCS*, pages 265–272.